\documentclass[sigconf,nonacm]{acmart}
\usepackage{amsfonts}
\usepackage{array}
\usepackage{placeins}
\usepackage{subcaption}
\setcopyright{none}
\renewcommand{\footnotetextcopyrightpermission}[1]{}

\begin{document}

\title{Multivariate Scientific Data Compression with Learned Cross-Variable Latent Decorrelation and Autoregressive Entropy Modeling}

\settopmatter{authorsperrow=3}

\author{Liangji Zhu}
\affiliation{
  \department{Department of Computer \& Information Science \& Engineering}
  \institution{University of Florida}
  \city{Gainesville}
  \state{FL}
  \country{USA}
}
\email{zhu.liangji@ufl.edu}

\author{Anand Rangarajan}
\affiliation{
  \department{Department of Computer \& Information Science \& Engineering}
  \institution{University of Florida}
  \city{Gainesville}
  \state{FL}
  \country{USA}
}
\email{anand@cise.ufl.edu}

\author{Sanjay Ranka}
\affiliation{
  \department{Department of Computer \& Information Science \& Engineering}
  \institution{University of Florida}
  \city{Gainesville}
  \state{FL}
  \country{USA}
}
\email{ranka@cise.ufl.edu}

\begin{abstract}
Scientific simulations generate collections of physical fields with heterogeneous statistics and dependencies, yet learned compressors often encode those fields independently or rely on a shared encoder without explicitly modeling the structure that remains in latent space. We present CAESAR-LDAR, an error-controlled multivariate learned compressor that augments a shared CAESAR-V backbone with two complementary mechanisms: a trainable orthogonal transform that reorganizes dependence across aligned latent channels, and a causal autoregressive hierarchical prior that captures local spatial structure left after transformation. Orthogonality is maintained through a matrix-exponential parameterization, making the transform exactly invertible without an additional penalty. A common residual-correction stage is applied uniformly to all variants to enforce the requested reconstruction tolerance.

Experiments across combustion, climate, and turbulence data show that the two mechanisms are useful in different regimes. Latent decorrelation helps most when substantial linear cross-channel dependence survives the nonlinear encoder, whereas autoregressive modeling remains effective when the remaining structure is primarily local or spatial. Their combination provides the strongest or near-strongest rate–distortion performance across the evaluated datasets. The global transform adds little computational overhead, while autoregressive coding introduces a larger throughput tradeoff. More broadly, the results suggest a practical design principle for multivariate scientific compression: exploit global cross-channel dependence when it is measurably present in latent space, and use local probabilistic context as a complementary mechanism across a wider range of data regimes.

\end{abstract}


\maketitle

\section{Introduction}
Large simulations increasingly produce collections of aligned fields that are expensive to store and move~\cite{10.1145/3733104,CAPPELLO2025107323}. Most scientific compressors exploit redundancy within each field--spatial smoothness, temporal coherence, or local predictability--but do not explicitly model dependence that remains across aligned variables after encoding. SZ~\cite{SZ} and ZFP~\cite{doi:10.1137/19M126904X} provide efficient error-controlled coding, while learned codecs adapt their representations to a scientific domain. CAESAR-V~\cite{app15168977}, for example, shares one model across variables, yet its entropy model largely treats the encoded channels independently. Our central premise is that a shared nonlinear encoder may leave structured cross-channel dependence in latent space, and that explicitly modeling this residual dependence can reduce the coded rate without duplicating the backbone.

We keep the CAESAR-V analysis and synthesis networks and add two modules between the encoder and the entropy coder. The shared nonlinear encoder first maps each channel to a compact latent representation of recurring spatiotemporal structure. Latent decorrelation (LD) then centers aligned latents and applies a learned orthogonal transform, changing the basis in which they are quantized and coded without changing the encoder input or requiring a separate network for each variable group. Autoregressive modeling (AR) replaces the scale-only prior with a Minnen-style hierarchical context model~\cite{Minnen} that uses already decoded neighbors within each two-dimensional latent plane. The two modules therefore address complementary structure: LD acts across aligned attributes after nonlinear feature extraction, while AR captures local spatial dependence within the resulting components.

For S3D and E3SM, each aligned channel is a physical variable. JHTDB has an additional spatial dimension, so we retain 20 native planes and represent its three physical fields as $3\times20=60$ aligned variable--plane channels. This layout presents the native 3D data to the shared backbone; consequently, JHTDB measures aligned cross-channel dependence that may include both cross-variable and cross-plane structure rather than a pure cross-variable effect.

Error control is supplied by the GAE and LBRC residual coders from our companion eScience study~\cite{lbrc2026}. At each operating point, we add the smaller correction stream that satisfies the requested tolerance. This stage is held fixed across CAESAR-V, LD, AR, and LDAR and operates only after learned reconstruction; it does not define the latent representation or probability model studied here. Conceptually, the learned path captures recurring macro-scale structure across space, time, and attributes, while the residual stream records the fine-scale, sample-specific deviations needed to meet the prescribed tolerance. Every reported total includes the learned bitstream, model and side information, and the selected correction stream. The present experiments use GAE/LBRC, but the LD and AR modules are not architecturally tied to either correction method.

The main question is not whether an orthogonal transform can lower correlation; it is when that reduction translates into fewer bits. We therefore measure the dependence that remains after nonlinear encoding and compare it with the observed rate savings. S3D, E3SM, and JHTDB are deliberately treated as three application cases rather than a controlled study of channel count. They differ in physics, dimensionality, and channel construction. What they provide is a useful range of operating regimes: strong LD gains on S3D, an AR-dominated result on E3SM, and smaller gains from both components on JHTDB.

The contribution is a systematic architecture for explicitly modeling residual latent dependence after a shared nonlinear encoder, together with evidence about when each mechanism pays off. We introduce an invertible aligned-channel transform that preserves one shared backbone, combine it with an autoregressive hierarchical prior for local latent context, and separate their contributions through ablations on three scientific datasets. We also compare against variable clustering and a fixed global PCA/KLT basis. Neither orthogonal transforms nor autoregressive priors are new by themselves. Their placement after a shared nonlinear encoder reveals which form of redundancy remains available for coding.

Placing the transform after the shared encoder is central to this design. Early fusion would tie the first network layer to a fixed set and ordering of variables, while separate multivariate models would give up the storage advantage of one shared backbone. The latent-space construction leaves the per-channel analysis transform unchanged and only couples aligned representations after feature extraction. It therefore preserves the original CAESAR-V interface, permits the same convolutional weights to be reused across all channels, and confines the added multivariate state to one mean vector and one global matrix. Because the transform is full rank, it does not discard components or create a second approximation stage; any coding gain must come from a basis that works better with quantization and probability modeling.

\section{Related Work}
\paragraph*{Scientific data compression.}
Traditional compressors combine block transforms, prediction, tensor decompositions, or multiresolution representations. ZFP uses near-orthogonal block transforms~\cite{doi:10.1137/19M126904X}; SZ combines prediction, quantization, and entropy coding~\cite{SZ}; tensor approaches identify multilinear low-rank structure~\cite{8663447}; and multilevel schemes encode corrections across scales~\cite{GONG2023101590,10.1145/3712285.3759878}. Predictive pipelines such as DPCM and interpolation-based coding exploit local smoothness~\cite{DCPM,ACM}, while FAZ combines wavelet processing with adaptive prediction~\cite{FAZ}. These methods are mature, fast, and require no dataset-specific training, but most operate on each physical variable separately and therefore cannot directly exploit aligned cross-variable dependence.

\paragraph*{Learned compression and entropy models.}
Learned codecs based on variational autoencoders jointly optimize analysis, synthesis, and probability models~\cite{AE}. Ball\'e et al. introduced scale hyperpriors that transmit side information for spatially varying latent distributions~\cite{Balle}. Minnen et al. combined hierarchical and autoregressive context, conditioning each latent symbol on both the hyperprior and previously decoded neighbors~\cite{Minnen,lee2019contextadaptiveentropymodelendtoend}. Scientific adaptations use 3D convolutions, super-resolution, or attention to model spatiotemporal structure~\cite{app15168977,li2025generative,Feng_2025_CVPR}. Coordinate-based neural representations provide another route for multidimensional weather and climate data~\cite{DBLP:conf/iclr/HuangH23}. CAESAR-V provides the shared backbone used here, but its standard formulation processes variables independently.

\paragraph*{Multivariate and decorrelating methods.}
Cross-field redundancy has been modeled through distributed Tucker decomposition~\cite{7516088}, multilevel multivariate reduction~\cite{ainsworth2019multivariate}, transfer across related variables~\cite{wang2022locality}, and joint learned representations~\cite{han2025dualstage}. Existing approaches commonly mix variables at the input, learn a monolithic joint representation, or train specialized models for subsets of fields. Our approach instead preserves a shared per-variable nonlinear encoder and introduces one global orthogonal transform only after feature extraction. This retains the original encoder input structure, permits any number of aligned variables to share the same backbone, and avoids the model-storage cost of maintaining a separate network for every variable group. Because the transform is optimized together with quantization and entropy modeling, it is not restricted to maximizing variance concentration as in classical PCA/KLT; it can organize the latent coordinates according to their eventual coding cost.

\paragraph*{Residual correction.}
Learned scientific codecs require a correction stream when a prescribed reconstruction tolerance must be met. Our companion eScience work~\cite{lbrc2026} develops complementary GAE and LBRC backends. GAE is generally effective at moderate tolerances, while LBRC becomes preferable in the high-fidelity regime. We use the lower valid total rate at each operating point and apply the same rule to every ablation. The correction stage is supporting infrastructure here; the present study concerns the learned representation and entropy model that precede it.

\section{Method}
\subsection{Problem and Architecture}
Let $\mathbf{x}$ contain $V$ aligned physical variables, and let $N_v$ be the number of samples in variable $v$, including any retained extra spatial dimension. Reconstruction must satisfy
\begin{equation}
\overline{\mathrm{NRMSE}}
=\frac{1}{V}\sum_{v=1}^{V}\frac{\|\mathbf{x}^{(v)}-\hat{\mathbf{x}}^{(v)}\|_2}{\sqrt{N_v}(\max\mathbf{x}^{(v)}-\min\mathbf{x}^{(v)})}\leq\tau.
\label{eq:nrmse}
\end{equation}
The shared CAESAR-V backbone processes $G$ aligned spatiotemporal input blocks. For S3D and E3SM, each block corresponds to one physical variable and $G=V$. When a native third spatial dimension of length $D$ is retained as a stack of planes, each block is a variable–plane pair and $G=VD$; JHTDB uses this representation. Each normalized block is passed through the same encoder.
At a fixed latent feature, time index, and in-plane location, the corresponding values from the $G$ encoded blocks form one aligned vector. LD rotates this vector before quantization. The decoder performs the inverse rotation, reconstructs each channel with the shared synthesis network, and reassembles the planes before Eq.~\eqref{eq:nrmse} is evaluated on the original physical variables.

The coding path is otherwise unchanged. The hyperlatent is transmitted first and provides spatially varying side information for the primary latent. AR variants also condition each symbol on previously decoded in-plane neighbors. rANS encodes the resulting discrete probabilities. After learned decoding, a correction stream is added only if the reconstruction does not meet the target. This order matters: LD changes the representation that is quantized, AR changes the probability assigned to the quantized symbols, and GAE/LBRC corrects any remaining reconstruction error. The modules therefore act at distinct points in the pipeline rather than duplicating the same operation.

\begin{figure*}[t]
\centering
\includegraphics[width=0.99\textwidth]{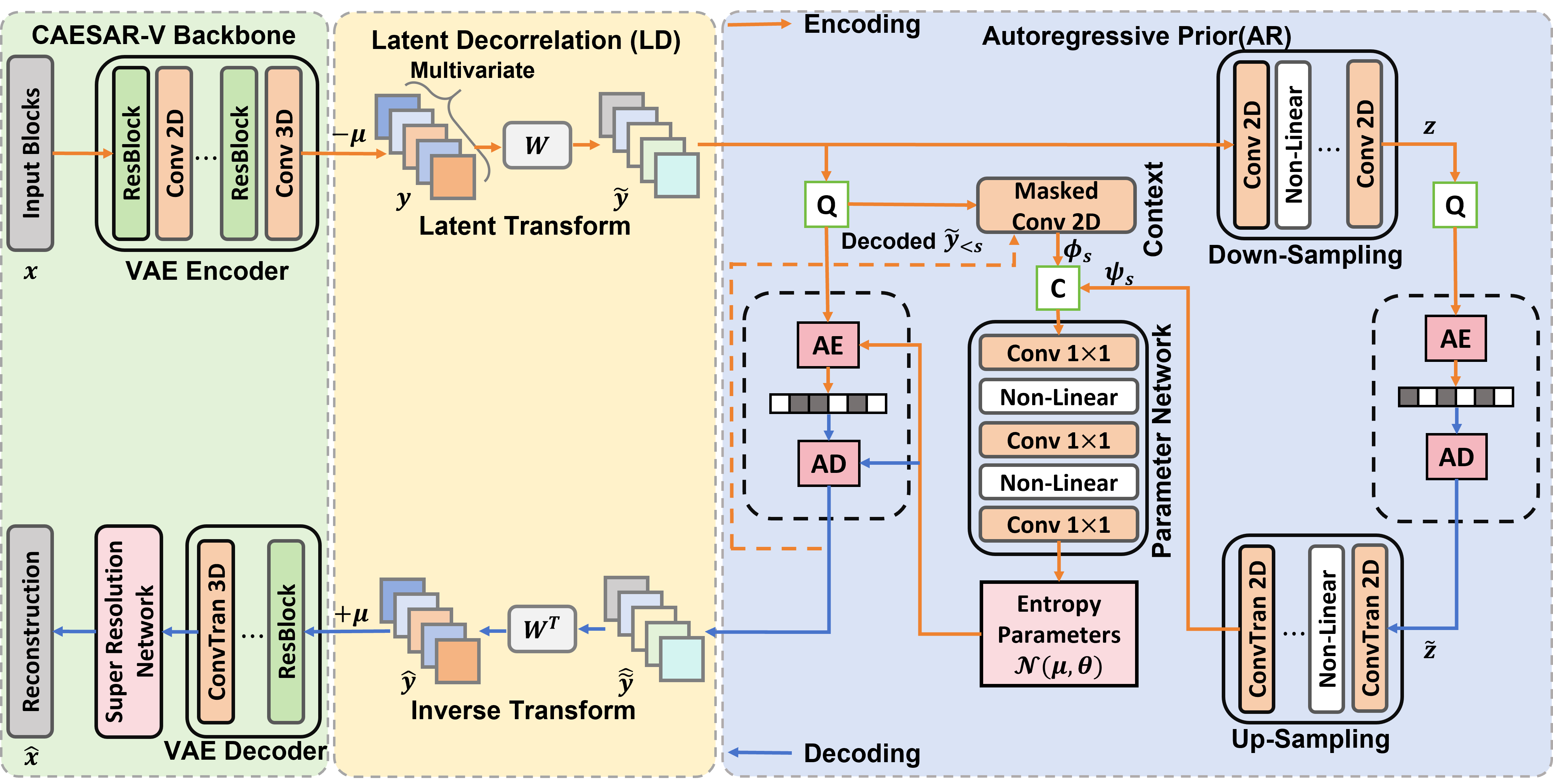}
\caption{CAESAR-LDAR. A shared CAESAR-V backbone encodes aligned channel blocks. The latent-decorrelation module centers the resulting latents and applies a learned orthogonal transform $\mathbf{W}$; $\mathbf{W}^{\top}$ applies the inverse transform. The entropy model combines hyperprior features with causal within-channel spatial context, where ``C'' denotes channel-wise concatenation of hyperprior and context features, ``Q'' denotes rounding-based quantization, and ``AE'' and ``AD'' denote arithmetic encoding and decoding, respectively. ``Conv 2D/3D'' denotes a stride-2 convolutional layer, whereas ``ConvTran 3D'' denotes a transposed-convolutional layer used for upsampling. After learned decoding, the lower-rate valid GAE/LBRC correction stream is appended to ensure error-bounded reconstruction.}
\label{fig:architecture}
\end{figure*}

\subsection{Aligned-Channel Latent Transform}
Let $\mathbf{y}\in\mathbb{R}^{G\times C\times T'\times H'\times W'}$ be the stacked latent tensor. We first compute the mean of each aligned channel,
\begin{equation}
\mu_g=\frac{1}{CT'H'W'}\sum_{c,t,h,w}y_{gcthw},
\end{equation}
and apply a learned matrix $\mathbf{W}\in\mathbb{R}^{G\times G}$ to the centered channel vector at every latent feature and location:
\begin{equation}
\widetilde{y}_{ucthw}=\sum_{g=1}^{G}W_{ug}(y_{gcthw}-\mu_g),\qquad \mathbf{W}^{\top}\mathbf{W}=\mathbf{I}.
\label{eq:transform}
\end{equation}
We enforce orthogonality with PyTorch's matrix-exponential parametrization rather than a penalty. Let $\mathbf{M}$ be unconstrained and $\mathbf{L}=\operatorname{tril}(\mathbf{M},-1)$. At each forward pass,
\begin{equation}
\mathbf{A}=\mathbf{L}-\mathbf{L}^{\top},\qquad
\mathbf{W}=\exp(\mathbf{A}),\qquad
\mathbf{y}=\mathbf{W}^{\top}\widetilde{\mathbf{y}}+\boldsymbol{\mu}.
\label{eq:orthogonal_param}
\end{equation}
Because $\mathbf{A}$ is skew-symmetric, $\exp(\mathbf{A})$ is orthogonal up to floating-point precision. Adam optimizes $\mathbf{M}$ through the differentiable matrix exponential; no orthogonality penalty is required.
Alternative orthogonality-preserving options include Stiefel-manifold optimization~\cite{li2025stella} and Cayley transforms; we use matrix-exponential parametrization because it supports standard unconstrained optimization at the moderate transform sizes considered here.
Setting $\mathbf{M}=\mathbf{0}$ gives $\mathbf{W}=\mathbf{I}$, so training starts from CAESAR-V.
The same $\mathbf{W}$ is used for every latent feature, time index, and in-plane location. The model stores one $G\times G$ matrix, while the $G$ block means are sent once per multivariate sample. Applying the transform costs a dense matrix--vector product for each aligned latent vector, or $O(G^2)$ arithmetic, but no block-specific basis bits. In the tested models this work is small relative to the convolutional encoder and decoder, which is consistent with the throughput results in Section~\ref{sec:runtime}.

We chose a global basis rather than predicting a new transform for every block. A block-adaptive basis could capture local changes in coupling, but it would also require either additional side information or a conditioning network that must be reproduced at the decoder. The global transform is intentionally narrower: it asks whether a single, reusable change of basis can expose dependence that the shared encoder leaves behind. This makes the added storage easy to account for and keeps the LD path fully parallel. It also sets a clear limitation--dependence that varies strongly with location or state may not be captured by one matrix.

Orthogonality gives the comparison a useful interpretation. Before quantization, the transform preserves the Euclidean norm of every aligned latent vector; it does not reduce distortion by shrinking the representation or by dropping low-variance directions. Its role is to redistribute the coordinates so that the scalar quantizer and entropy model see a more favorable collection of marginals. The transform may concentrate predictable structure into fewer components, but all $G$ components are still coded and inverted. This is different from dimensionality reduction, where omitted components trade rate directly for transform error. It also explains why a lower Pearson correlation is supporting evidence rather than the optimization target: the model is trained on reconstruction error and estimated rate, and it is accepted only when the measured bitstream becomes smaller at matched quality.

Fixed-PCA CAESAR-V provides the closest classical comparison. We freeze a pretrained CAESAR-V encoder, decoder, and super-resolution module, estimate one PCA/KLT mean and basis from training latents, retain all components, and then retrain only the hyper-encoder, hyper-decoder, and prior. The PCA basis is stored once and reused for every block. Because no component is discarded, the baseline adds no truncation error. This is a practical codec comparison, not a pure test of one matrix against another: Fixed-PCA uses a representation learned before the basis is fitted, whereas CAESAR-LD lets the encoder, transform, quantizer, and entropy model adapt together.

\subsection{Autoregressive Entropy Model and Training}
The baseline scale hyperprior models quantized latents $\hat{\mathbf{y}}$ conditioned on hyperlatents $\hat{\mathbf{z}}$. The hyper-decoder predicts spatially varying Gaussian scales, while the hyperlatent itself is coded with a fully factorized prior. Its estimated rate is $R_y+R_z$, where $R_y=\mathbb{E}[-\log_2 p(\hat{\mathbf{y}}\mid\hat{\mathbf{z}})]$ and $R_z=\mathbb{E}[-\log_2 p(\hat{\mathbf{z}})]$. We augment this model with a causal $5\times5$ masked convolution. For spatial position $s$ and latent channel $c$,
\begin{equation}
p(\hat{y}_{c,s}\mid\hat{\mathbf{Y}}_{<s},\hat{\mathbf{z}})=\left[\mathcal{N}(\mu_{c,s},\sigma_{c,s}^2)*\mathcal{U}(-\tfrac12,\tfrac12)\right](\hat{y}_{c,s}),
\label{eq:ar}
\end{equation}
where $(\mu_{c,s},\sigma_{c,s})$ are predicted from the hyperprior and causal context. If $\mathbf{h}_s$ is the hyper-decoder feature and $\boldsymbol{\phi}_s$ is the masked-convolution feature, the entropy-parameter network computes $[\boldsymbol{\mu}_s;\boldsymbol{\sigma}_s]=f_{\mathrm{ep}}([\mathbf{h}_s;\boldsymbol{\phi}_s])$. The mask excludes the current symbol, positions to its right in the same row, and all later rows. Encoder and decoder therefore see exactly the same context.

The hyperlatent is decoded before the primary latent. The codec then visits the two-dimensional latent grid in raster order. At each position it predicts the Gaussian parameters for all feature channels, encodes or decodes the corresponding symbols with rANS, and inserts the reconstructed values into the context buffer. Different transformed components use the same network parameters but separate causal contexts. Cross-channel dependence is handled by $\mathbf{W}$ rather than by conditioning one transformed component directly on another. During training the masked convolution can be evaluated in parallel; actual entropy coding remains sequential. This explains why LD has almost no measured runtime cost while AR is substantially slower.

For S3D and E3SM, this division is literal: LD acts across variables and AR acts within each variable's spatial plane. For JHTDB, the aligned dimension also contains plane identity. LD can therefore reorganize both cross-variable and cross-plane structure, while AR models local context inside each retained plane. We keep this distinction explicit when interpreting the JHTDB results.

The model is trained end-to-end with
\begin{equation}
\mathcal{L}=\mathrm{MSE}(\mathbf{x},\hat{\mathbf{x}})+\lambda(R_y+R_z).
\label{eq:loss}
\end{equation}
The training objective optimizes the learned stream; the error bound is enforced after decoding. Let $R_m(\tau)$ denote the correction rate produced by backend $m$ when its reconstruction satisfies Eq.~\eqref{eq:nrmse}. The reported rate is
\begin{equation}
R_{\mathrm{total}}(\tau)=R_{\mathrm{learned}}+\min_{m\in\{\mathrm{GAE},\mathrm{LBRC}\}\,:\,\mathrm{valid}}R_m(\tau),
\label{eq:correction_select}
\end{equation}
with a zero correction rate when the learned reconstruction already meets the target. The selection is made independently at each operating point and is applied identically to all four CAESAR variants. Every plotted point therefore satisfies the requested criterion by construction. Actual rANS streams, model parameters, transform and normalization metadata, and the selected correction stream are all included in the reported size.

\subsection{Bitstream and Reconstruction Procedure}
A compressed item contains the primary-latent and hyperlatent rANS streams together with the information needed to reproduce the learned transform and restore the physical fields. The trained network parameters and the global matrix $\mathbf{W}$ are charged to the model portion of the compressed representation. Per-sample information includes the latent mean vector, normalization parameters, tensor dimensions, and channel-to-variable layout. For JHTDB, the layout records how the 60 variable--plane channels are reassembled into three three-dimensional fields. These costs are included rather than reporting stream-only compression. Clustered CAESAR-V is likewise charged for every separately trained model.

Decoding follows a fixed order. The decoder first restores the metadata and hyperlatent, then reconstructs the primary latent using either the scale-only prior or the causal AR prior. LD variants apply $\mathbf{W}^{\top}$ and add the transmitted channel means before the shared synthesis network is run. The reconstructed channels are denormalized and, for JHTDB, the retained planes are placed back on the native third spatial axis. Only then is the selected GAE or LBRC correction decoded and added. The error check is performed on the reassembled physical variables, not on the intermediate channel representation. This is why JHTDB can use variable--plane channels internally while still reporting one NRMSE value per physical field.

The separation between learned coding and residual correction is also useful experimentally. LD and AR can be compared through their latent rates, model cost, and effect on the residual, while Eq.~\eqref{eq:correction_select} ensures that the final rate corresponds to a valid error-controlled reconstruction. A model that produces a smaller learned stream but a larger residual is not automatically favored; its correction cost is included before the total is reported. A better learned reconstruction can lower both parts of the bitstream. The same accounting is used in every curve and in Table~\ref{tab:rate}; further coding details are given in Appendix~\ref{app:coding}.

\section{Experiments}
\subsection{Setup}
Table~\ref{tab:data} summarizes the three evaluated regimes. For each variable, NRMSE is computed relative to its own value range, and the reported quality is the macro-average $V^{-1}\sum_v\mathrm{NRMSE}_v$. Compression ratio is original size divided by total compressed size. S3D uses float64; E3SM and JHTDB use float32. Reported sizes include all streams, model parameters, latent means, transform and normalization information, and the selected correction stream. SZ3 and ZFP compress each variable independently and are compared at matched achieved NRMSE.

\begin{table}[t]
\centering
\caption{Datasets and evaluated tensors.}
\label{tab:data}
\small
\setlength{\tabcolsep}{3.5pt}
\begin{tabular}{lcccc}
\toprule
Data & Phys. vars. & Chans $(G)$ & Frames & Spatial grid \\
\midrule
S3D~\cite{YOO20111727} & 58 & 58 & 64 & $640^2$ \\
E3SM~\cite{https://doi.org/10.1029/2018MS001603} & 5 & 5 & 720 & $240^2$ \\
JHTDB~\cite{li2008jhtdb} & 3 & 60 & 240 & $512^2$ \\
\bottomrule
\end{tabular}
\end{table}

S3D models compression ignition using 58 coupled chemical species; we evaluate 64 frames on a $640\times640$ grid. E3SM contains five aligned but more heterogeneous climate variables over 720 frames on a $240\times240$ grid. JHTDB contains two Cartesian velocity components, $u_x$ and $u_y$, and pressure $p$ on a 3D domain. Each evaluated tensor retains 20 native planes and a $512\times512$ in-plane grid, yielding $G=3\times20=60$ variable--plane channels. Thus, the three datasets represent different dependence regimes rather than a controlled channel-count study; Appendix~\ref{app:jhtdb} gives the JHTDB construction and reassembly details.

We compare CAESAR-V, CAESAR-LD, CAESAR-AR, and CAESAR-LDAR with SZ3 and ZFP. The latter are run through their Python interfaces with parameters swept to generate rate--distortion curves. Each traditional compressor processes variables independently, and all methods are compared at the achieved macro-NRMSE rather than at nominal internal error settings. This distinction matters because internal tolerance parameters do not map identically to the macro-NRMSE used here. We therefore read each curve at matched achieved quality and avoid treating a nominal compressor setting as an equivalent operating point.

All learned variants within a dataset use the same train/test split, normalized inputs, CAESAR-V backbone, and optimization budget. The ablations change only the LD and AR modules. Model-inclusive accounting is used throughout: CAESAR-V is charged once for its shared network, the clustered baseline is charged for each cluster-specific network, and LD variants include the transform and its side information. The Fixed-PCA point likewise includes the stored mean and basis. This makes the comparisons sensitive to the practical cost of additional specialization rather than only to the size of the entropy-coded latent stream.

Models are implemented in PyTorch 2.7 and trained on one NVIDIA B200 GPU for 100,000 iterations with Adam. Samples contain 16 consecutive frames cropped to $256\times256$; multivariate variants use aligned blocks, and JHTDB retains its 20 aligned planes. The learning rate begins at $4\times10^{-4}$ and is halved at 20\%, 40\%, 60\%, and 80\% of training. The rate--distortion weight is $10^{-5}$ for the first 50,000 iterations and $2\times10^{-5}$ thereafter. Within each dataset, all variants use the same backbone, sampling, optimization schedule, and training budget, and the trained model is included in the compressed representation. All curves include primary and hyperlatent streams, model and transform information, normalization metadata, tensor-layout metadata, and the lower valid GAE/LBRC correction stream.

\subsection{Variable Grouping and Fixed PCA}
S3D's 58 chemical fields provide a direct test of how information should be shared across variables. A shared CAESAR-V model minimizes model overhead but ignores variable identity. We therefore compare it with clustered variants that train one model per group, a fixed global PCA/KLT transform, and the learned transform. The clusters are formed from coarse spatiotemporal and statistical descriptors of each variable; Appendix~\ref{app:baselines} gives the baseline construction details.

\begin{figure}[t]
\centering
\includegraphics[width=0.92\columnwidth]{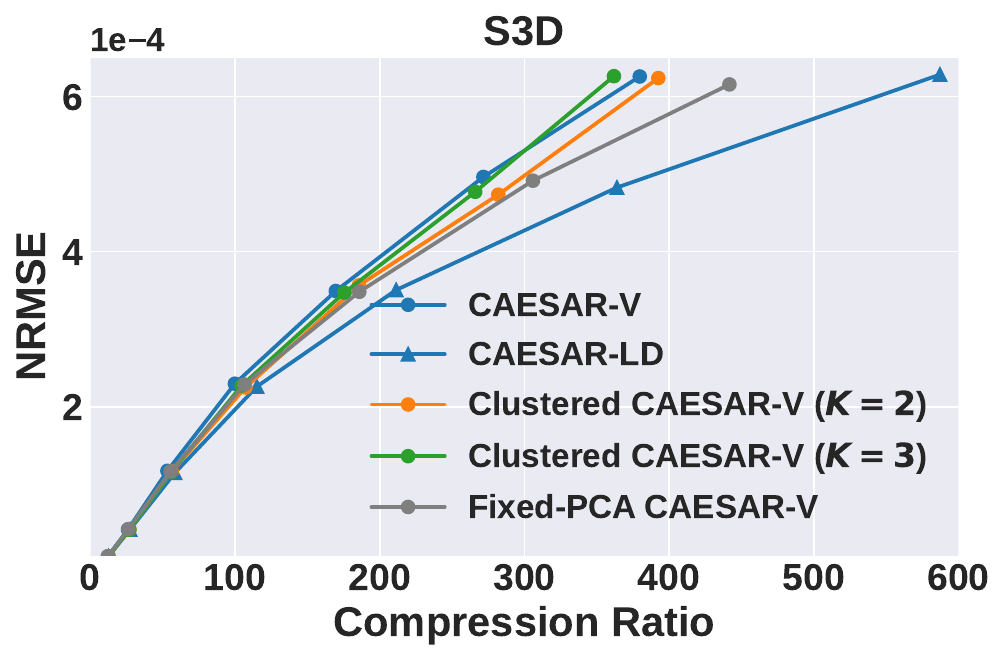}
\caption{S3D variable-aware baselines. Clustering and Fixed-PCA improve over shared CAESAR-V, while learned latent decorrelation performs best.}
\label{fig:grouping}
\end{figure}

Figure~\ref{fig:grouping} shows that two clusters provide a gain of up to about 5\% over the shared model at matched quality, confirming that variable-specific structure is useful. Increasing to three clusters does not help after the additional model storage is included, illustrating the trade-off between specialization and parameter overhead. In contrast, CAESAR-LD improves compression ratio by roughly 24--53\% over the shared model across the displayed range while retaining one shared backbone. Fixed-PCA consistently improves over shared CAESAR-V, confirming exploitable linear dependence, yet remains below CAESAR-LD. The ordering is important: clustering shows that field identity matters, PCA shows that a global linear basis can exploit part of that structure, and CAESAR-LD shows that the largest gain is obtained when the encoder, basis, quantizer, and entropy model adapt jointly. The learned method therefore avoids both the parameter duplication of clustering and the representation mismatch of a transform fitted after the encoder has already been trained.

\begin{figure*}[t]
\centering
\begin{subfigure}[t]{0.32\textwidth}
    \centering
    \includegraphics[width=\linewidth]{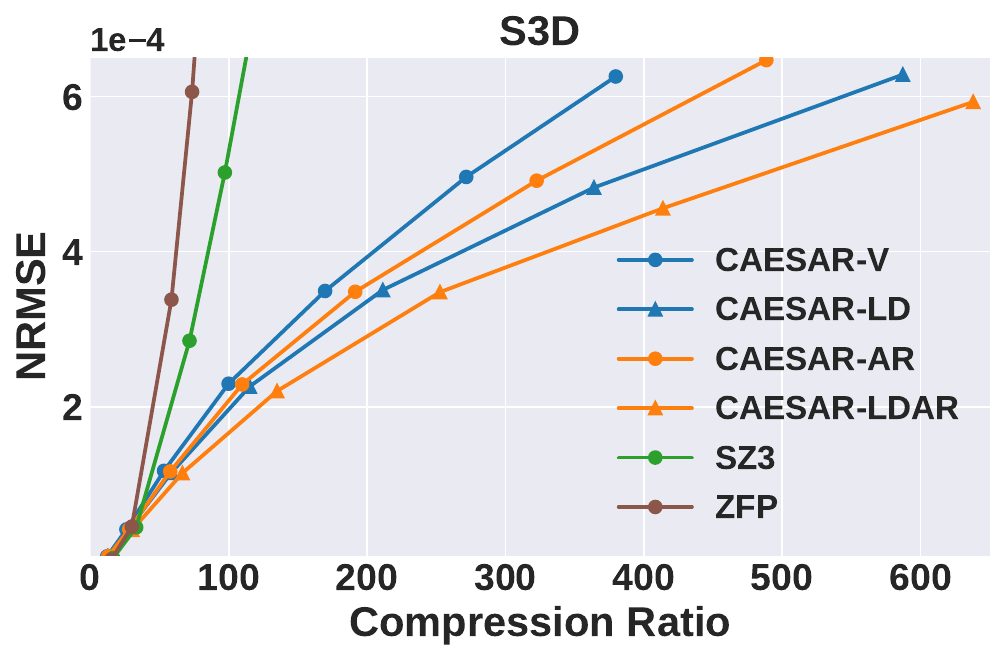}
    \caption{S3D}
    \label{fig:s3d}
\end{subfigure}
\hfill
\begin{subfigure}[t]{0.32\textwidth}
    \centering
    \includegraphics[width=\linewidth]{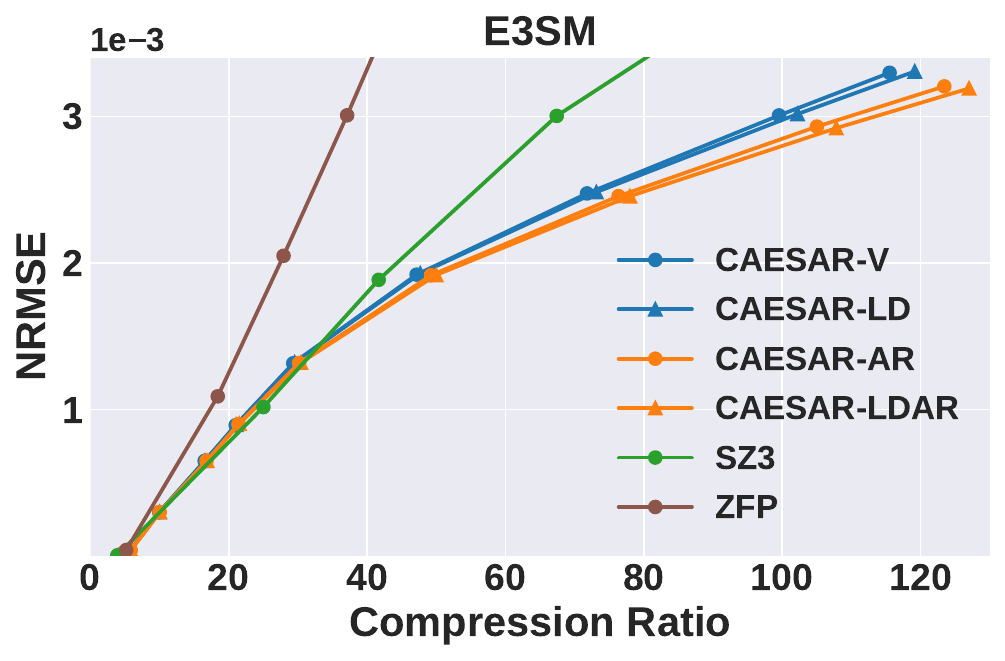}
    \caption{E3SM}
    \label{fig:e3sm}
\end{subfigure}
\hfill
\begin{subfigure}[t]{0.32\textwidth}
    \centering
    \includegraphics[width=\linewidth]{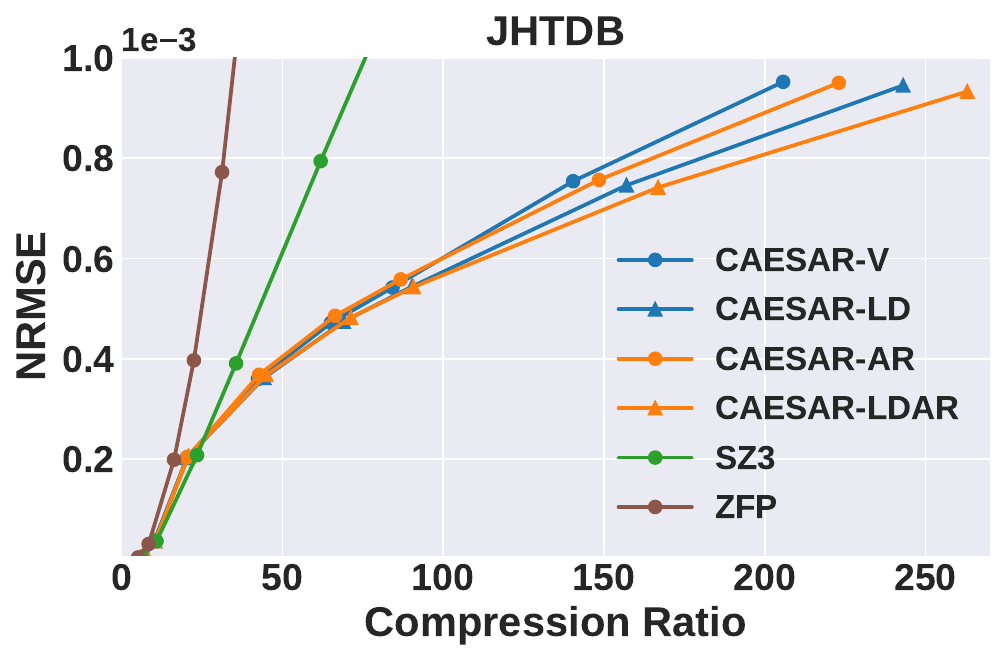}
    \caption{JHTDB}
    \label{fig:jhtdb}
\end{subfigure}
\caption{Rate--distortion performance across the three evaluated datasets. On S3D, both latent decorrelation (LD) and autoregressive entropy modeling (AR) improve CAESAR-V, and their combination provides the strongest performance. On E3SM, the improvement is primarily AR-driven. On JHTDB, the model uses 60 variable--plane channels from native 3D blocks, and CAESAR variants are evaluated with the lower-rate valid GAE or LBRC correction stream included at each operating point.}
\label{fig:rd_all}
\end{figure*}

\subsection{S3D Component Ablation}
Figure~\ref{fig:s3d} isolates latent decorrelation (LD) and autoregressive entropy modeling (AR). At $5\times10^{-4}$ NRMSE, CAESAR-LD and CAESAR-AR improve compression ratio over CAESAR-V by 44.1\% and 18.9\%, respectively, while CAESAR-LDAR improves it by 75.4\%. The complete model is approximately $5.0\times$ SZ3 and $7.1\times$ ZFP at this operating point. The larger LD gain confirms substantial exploitable dependence among S3D's 58 fields, while AR provides complementary savings after transformation. The combined improvement is larger than either isolated gain because the two components act at different stages: $\mathbf{W}$ first reorganizes the cross-variable representation, and the context model then sharpens symbol probabilities within each transformed component.


Table~\ref{tab:rate} shows where the gain originates. LD cuts the primary-latent rate by 51.3\%; AR reduces latent and hyperlatent rates by 24.8\% and 41.0\%. Their combination produces a learned stream nearly as small as that of CAESAR-LD while requiring substantially less residual correction. The total falls from 0.232 to 0.132 bits/value, including model, side information, and GAE correction at this moderate-error point. Although AR increases model storage, the reduction in coded latents more than offsets that cost. The gain narrows at tighter tolerances because residual correction becomes a larger fraction of the total, but every displayed point uses the better valid GAE/LBRC result rather than a fixed switching rule.

The decomposition also separates two effects that are hidden by the headline compression ratio. Excluding the correction stream, CAESAR-V uses about 0.1305 bits/value, whereas CAESAR-LD uses 0.0778 and CAESAR-LDAR uses 0.0783. LD therefore accounts for most of the reduction in the learned representation. The full model nevertheless has the lowest total rate because its correction stream is only 0.0540 bits/value, compared with 0.0833 for LD and 0.1014 for CAESAR-V. In other words, joint LDAR training gives up a negligible amount in learned-stream size relative to LD alone and recovers much more through a smaller error-correction payload. The additional AR parameters cost roughly 0.0056 bits/value at this operating point, while the transform side information changes the side rate by only about $4\times10^{-5}$ bits/value. These values explain why LD is the attractive low-overhead option and why the complete model remains best when rate is the main objective.

\subsection{Latent Dependence}
\begin{figure}[htb]
\centering
\includegraphics[width=\columnwidth]{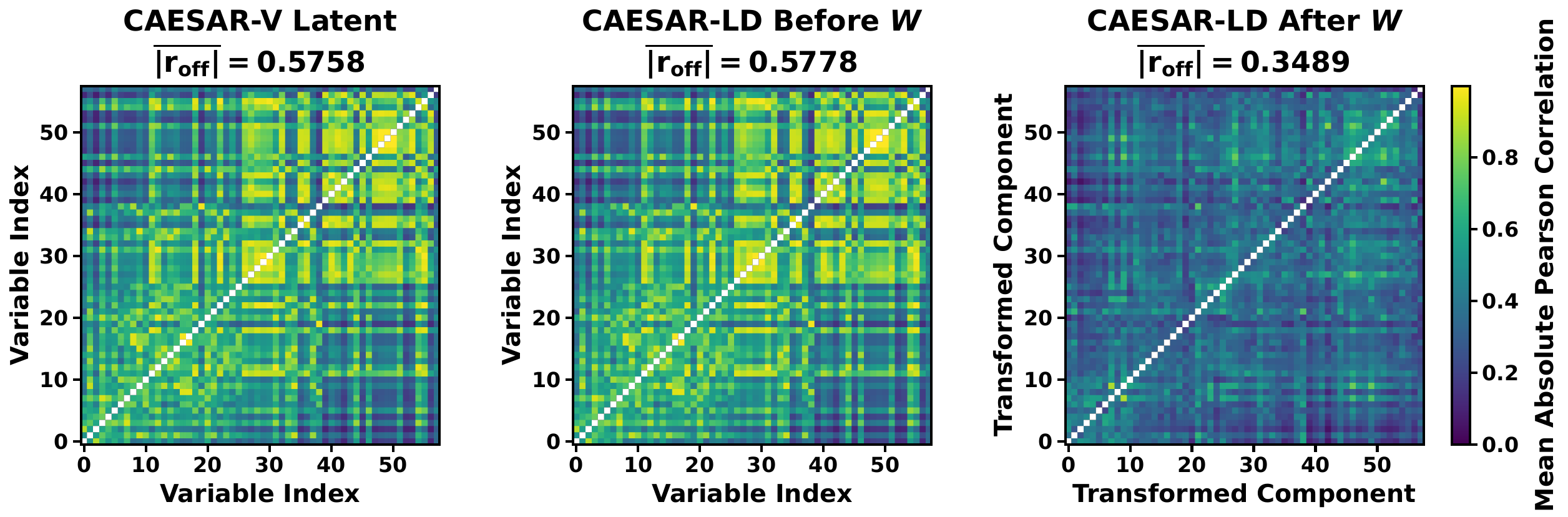}
\caption{S3D mean absolute latent-correlation matrices before quantization. The learned transform broadly reduces cross-component dependence.}
\label{fig:dependence}
\end{figure}

For each test block, continuous pre-quantization latents are reshaped to $G\times M$, and we compute the mean absolute off-diagonal Pearson correlation $\rho$. On S3D, CAESAR-V and the pre-transform CAESAR-LD representation have nearly identical dependence ($0.5758$ and $0.5778$), while the learned transform reduces it to $0.3489$--a 39.6\% reduction--together with the latent-rate reduction in Table~\ref{tab:rate}. Figure~\ref{fig:dependence} shows that the reduction is distributed across the matrix rather than confined to a few component pairs.

The near-equality of the CAESAR-V and pre-transform CAESAR-LD correlations indicates that retraining the shared encoder alone does not explain the change; the decrease occurs when $\mathbf{W}$ is applied. The corresponding primary-latent rate drops from 0.0952 to 0.0464 bits/value, linking the measured dependence reduction with actual coding savings. Because the transform is global and shared by all blocks, this gain is obtained without transmitting a block-specific basis.

Throughout, decorrelation means reduction in linear second-order dependence; we do not claim that $\mathbf{W}$ removes all nonlinear dependence. This is nevertheless the dependence that the module is designed to address. Scalar quantization and the marginal part of the entropy model operate along latent coordinate axes, so a learned rotation can present those components in a basis that is easier to code even when higher-order dependence remains. Pearson correlation is therefore explanatory evidence, not the objective or proof of compression; the actual benefit is established by the measured bitstream. Both the dependence that survives the encoder and the absolute amount removed by $\mathbf{W}$ matter, since relative reduction alone can hide large differences in starting correlation.

\begin{table*}[t]
\centering
\caption{S3D rate decomposition at $5\times10^{-4}$ NRMSE (bits/value).}
\label{tab:rate}
\small
\setlength{\tabcolsep}{5pt}
\begin{tabular}{lccccccc}
\toprule
Method & Latent & Hyper & Side & Model & Correction & Total & CR \\
\midrule
CAESAR-V & .0952 & .0194 & .00009 & .0158 & .1014 & .2320 & 275.9 \\
CAESAR-LD & .0464 & .0154 & .00013 & .0159 & .0833 & .1611 & 397.3 \\
CAESAR-AR & .0716 & .0115 & .00009 & .0213 & .0906 & .1950 & 328.2 \\
CAESAR-LDAR & .0471 & .0097 & .00013 & .0214 & .0540 & \textbf{.1323} & \textbf{483.8} \\
\bottomrule
\end{tabular}
\end{table*}


The baseline and pre-transform values are nearly identical on all three datasets, so retraining the shared encoder alone does not explain the lower correlations; the change occurs after $\mathbf{W}$. The starting value and absolute decrease distinguish the regimes. S3D begins at 0.5778 and removes 0.2289, whereas JHTDB begins at 0.3061 and removes 0.1189. Although both reductions are about 39\% in relative terms, S3D removes nearly twice as much absolute dependence and obtains the larger LD gain. E3SM begins at only 0.1868 and changes by 0.0053, leaving little globally linear dependence for the transform to exploit. Thus, relative reduction alone is insufficient: we report pre-transform correlation, post-transform correlation, and absolute decrease and verify the interpretation against the actual coded rate.

S3D and JHTDB use transforms of similar size, but their channels have different meanings: chemical species for S3D and three fields across 20 native planes for JHTDB. Their comparison therefore rules out a simple channel-count explanation but is not a controlled dimensionality experiment. Pearson correlation is an interpretable first-order diagnostic rather than a direct measure of entropy; the compression claim rests on the measured bitstream.



The datasets expose distinct regimes. S3D contains many tightly coupled fields and shows a large gain from LD. On E3SM (Figure~\ref{fig:e3sm}), CAESAR-AR consistently improves over CAESAR-V, whereas LD adds only a small increment; its moderate overall improvement is therefore driven primarily by autoregressive modeling. The low pre-transform correlation and $0.0053$ absolute decrease indicate that little globally linear cross-variable structure remains after encoding, while useful spatial context persists. JHTDB (Figure~\ref{fig:jhtdb}) shows smaller but consistent improvements, with CAESAR-LDAR strongest over most of the range. Its pre-transform correlation is 0.3061 and the absolute decrease is 0.1189, both below S3D. Because JHTDB's 60 channels include the native plane axis, its LD gain may combine cross-variable and cross-plane redundancy and is interpreted as aligned-channel, not purely cross-variable, evidence.

\begin{table}[H]
\centering
\caption{Mean absolute off-diagonal latent correlation. JHTDB's 60 channels are three physical variables across 20 retained planes of its native third spatial dimension.}
\label{tab:dependence}
\setlength{\tabcolsep}{2.5pt}
\begin{tabular}{lccccc}
\toprule
Data &  chans. $(G)$ & Base & Pre-$W$ & Post-$W$ & $\Delta\rho$ \\
\midrule
S3D & 58 & 0.5758 & 0.5778 & 0.3489 & 0.2289 \\
E3SM & 5 & 0.1824 & 0.1868 & 0.1815 & 0.0053 \\
JHTDB & 60 & 0.3059 & 0.3061 & 0.1872 & 0.1189 \\
\bottomrule
\end{tabular}
\end{table}

Across the two non-S3D datasets, CAESAR-LDAR is the strongest learned configuration over most of the evaluated range, but the relative contributions differ: E3SM is principally an entropy-modeling result, whereas the JHTDB gain is modest for both additions. Together with the dependence measurements in Table~\ref{tab:dependence}, these results support an empirical regime characterization rather than a universal scaling law. LD is most valuable when substantial aligned dependence remains for the transform to remove; AR is useful more broadly because local conditional structure can persist even when global cross-variable decorrelation offers little additional benefit.

\subsection{Learned-Codec Runtime}
\label{sec:runtime}
Table~\ref{tab:runtime} reports component-level throughput of the learned codec on S3D. LD adds about 1\% overhead, whereas AR reduces throughput by roughly 60\% because causal entropy coding is sequential. LDAR closely matches AR, showing that the combined model adds negligible cost beyond the autoregressive prior. These values are not end-to-end error-bounded throughput and should not be compared directly with complete SZ3 or ZFP throughput: GAE/LBRC execution is an operating-point-dependent backend excluded from this table, although its full encoded size is included in every reported rate and compression ratio.

\begin{table}[H]
\centering
\caption{S3D learned-codec throughput (MB/s).}
\label{tab:runtime}
\small
\begin{tabular}{lrr}
\toprule
Method & Encode & Decode \\
\midrule
CAESAR-V & 404.1 & 301.4 \\
CAESAR-LD & 401.3 & 297.6 \\
CAESAR-AR & 163.0 & 122.3 \\
CAESAR-LDAR & 162.7 & 121.4 \\
\bottomrule
\end{tabular}
\end{table}

\section{Discussion}
The three datasets show why LD and AR should be kept separate. S3D retains strong dependence across many chemically related fields after the shared encoder, and a global change of basis removes a large fraction of the primary-latent rate. E3SM behaves differently. Its variables are physically connected, but their scales and spatial statistics are heterogeneous enough that the encoder leaves little common linear structure. The transform barely changes the measured correlation, while AR still benefits from local spatial predictability. JHTDB lies between these cases. Its starting correlation and absolute decrease are both lower than S3D's, and the measured rate gain is correspondingly smaller. Because its aligned dimension includes the third spatial axis, that gain should be read as cross-channel rather than purely cross-variable.

The comparison also cautions against reading physical coupling directly from the latent statistics. Variables can be coupled by governing equations without occupying a common linear subspace after a nonlinear encoder. Conversely, a transform can find useful statistical dependence even when the underlying variables have different physical meanings. The diagnostic in Table~\ref{tab:dependence} is therefore about the representation that is actually coded, not about physical coupling in the original data. We do not claim that the three datasets define a universal threshold for using LD. They show a consistent pattern that can be checked before committing to the added module: if the encoded channels start with little correlation and $\mathbf{W}$ cannot remove much in absolute terms, the expected benefit is small.

The cross-dataset comparison is deliberately descriptive rather than causal. The datasets differ in physics, precision, spatial dimensionality, number of frames, and the way channels are formed. They cannot isolate one factor in the way a controlled sweep over $G$ would. What they can establish is that the same architecture does not produce a uniform gain and that the observed differences are consistent with measurable properties of the encoded representation. S3D and JHTDB are especially useful together because they prevent a simple channel-count explanation: their transforms are similar in size, but their channel semantics and starting dependence differ. E3SM supplies the complementary low-dependence case. The resulting claim is therefore modest but operationally useful--measure the latent dependence left by the shared encoder before assuming that a cross-channel transform will pay for itself.

The Fixed-PCA result helps locate the source of the S3D improvement. A classical global basis already beats CAESAR-V, so some of the gain comes from ordinary linear dependence. At $5\times10^{-4}$ NRMSE, however, CAESAR-LD reaches a compression ratio of about 398 compared with about 315 for Fixed-PCA, a 26.3\% advantage. This comparison does not isolate the basis alone because the CAESAR-LD encoder adapts jointly while the PCA encoder is frozen. It answers the practical question instead: whether fitting PCA after training is enough. On S3D it is not. The nonlinear encoder learns a compact representation of the data manifold, and the full-rank $\mathbf{W}$ reorganizes that representation without reducing its dimension. Joint adaptation produces coordinates that work better with quantization and the learned entropy model.

The global nature of $\mathbf{W}$ is a deliberate engineering choice. It is stored once, requires no block-dependent signaling, and adds about one percent to learned-codec runtime. A conditioned or block-adaptive transform could model nonstationary coupling, but it would change both the computational cost and the bit accounting. The E3SM result marks the limitation of the present design rather than an inconsistency in the experiments: when a single global linear basis has little to remove, AR is the more useful addition. The present ablations are useful precisely because this low-cost module can be enabled or omitted independently.

The correction backend has a similarly specific role. GAE and LBRC make the reconstruction meet the requested tolerance; they are not used to create the LD or AR gain. Since the lower valid correction rate is selected for every variant, a better learned representation helps in two ways: it reduces the learned bitstream and can reduce the residual that must be corrected. This interaction is visible in Table~\ref{tab:rate}. At tighter tolerances the correction stream occupies a larger share of the total, so differences among the learned models naturally narrow. That behavior is expected for an error-controlled hybrid codec.

The dependence calculation can also guide model selection without training every configuration. The pre-transform statistic is available directly from continuous CAESAR-V training latents. A low starting value, as on E3SM, is an early warning that little linear redundancy remains for LD. When the starting value is appreciable, a pilot LD fit can check whether $\mathbf{W}$ produces a meaningful absolute decrease before committing to the full LDAR configuration. This is a heuristic rather than a universal threshold: the final choice should still use measured total rate, including correction, because correlation does not capture higher-order dependence or its interaction with quantization.

The measurements suggest two practical configurations. LD is attractive when the encoded channels retain appreciable aligned dependence: it is parallel, inexpensive, and on S3D captures most of the available gain at nearly baseline speed. AR is useful when additional rate savings justify sequential decoding, including E3SM where LD contributes little. LDAR gives the best rate over most of the tested range, but a workflow need not use every component. Archival storage may favor the full model, whereas checkpointing or interactive analysis may prefer LD alone. The shared backbone and correction protocol remain the same in either case.

\section{Conclusion and Limitations}
We added latent decorrelation and autoregressive entropy modeling to a shared CAESAR-V codec and evaluated them separately as well as together. The components do not help equally on every dataset. S3D is the favorable LD case: at $5\times10^{-4}$ NRMSE, CAESAR-LDAR reduces total rate by 43.0\% and raises compression ratio by 75.4\% over CAESAR-V. E3SM gains mainly from AR because little linear cross-variable dependence remains after encoding. JHTDB shows a smaller gain on a representation that maps three physical fields and 20 native planes to 60 aligned channels. Fixed-PCA improves on the baseline but remains below the jointly trained system.

The main conclusion is conditional. A global latent transform is worthwhile when the encoder leaves enough aligned dependence for the transform to remove; the starting correlation and its absolute decrease provide useful, though incomplete, evidence of that opportunity. AR applies more broadly, but its sequential coding order lowers throughput. The method also requires training for each dataset, and one global linear basis cannot capture arbitrary nonlinear or spatially varying relationships. JHTDB further shows that channel construction matters: its LD result mixes cross-variable and cross-plane structure and should not be interpreted as a controlled comparison with S3D. Finally, at very high fidelity the residual-correction stream becomes a larger part of the total size. These limits define the scope of the present result, while the ablations show that the two modules offer a practical rate--throughput choice within that scope.

More broadly, CAESAR-LDAR targets two major sources of redundancy: spatiotemporal patterns within fields and aligned relationships across attributes. The nonlinear backbone learns recurring macro-scale structure, LD and AR remove complementary redundancy in the latent representation, and residual correction restores the fine-scale information required by the error bound. The Fixed-PCA gap shows the value of joint learning over a linear basis fitted after the representation is fixed. These results suggest that explicitly modeling residual inter-variable or aligned-channel dependence is an important direction for learned compression as scientific datasets grow in variable count, dimensionality, and physical coupling.

\FloatBarrier
\clearpage
\bibliographystyle{ACM-Reference-Format}
\bibliography{references}

\clearpage
\appendix
\section{Additional Baseline Details}
\label{app:baselines}
\subsection{Clustered CAESAR-V}
The clustered baseline tests whether variable-specific specialization can recover some of the structure ignored by one shared CAESAR-V model. Each physical variable is represented by a descriptor containing both coarse spatiotemporal patterns and global statistics. The pattern component is formed by block-averaging the field, standardizing the coarse representation, and projecting it onto a low-dimensional PCA subspace. The statistical component contains the mean, standard deviation, value range, and normalized temporal and spatial finite-difference statistics.

Let $\mathbf{p}^{(v)}$ and $\mathbf{s}^{(v)}$ denote the pattern and standardized statistical features for variable $v$. After normalizing the two feature groups separately, the descriptor is
\begin{equation}
\mathbf{d}^{(v)}=
\left[
 w_p\frac{\mathbf{p}^{(v)}}{\alpha_p};\;
 w_s\frac{\mathbf{s}^{(v)}}{\alpha_s}
\right],
\end{equation}
where $\alpha_p$ and $\alpha_s$ are the root-mean-square magnitudes of the two groups and equal group weights are used. K-means partitions these descriptors, and a separate CAESAR-V model is trained for each cluster. Each block is still encoded independently by the model assigned to its physical variable; the baseline does not jointly encode aligned variables. All cluster-specific model parameters are included in the compressed-size accounting.

\subsection{Fixed-PCA CAESAR-V}
The Fixed-PCA baseline is designed to separate the value of a conventional global linear basis from end-to-end adaptation. We first train CAESAR-V normally. Its encoder, decoder, super-resolution module, and latent representation are then frozen. One global PCA/KLT mean and basis are estimated from training latents, all components are retained, and only the hyper-encoder, hyper-decoder, and entropy prior are retrained around the fixed representation. The stored mean and basis are included in the model and side-information cost. Because no component is discarded, this baseline introduces no transform truncation error. It should therefore be interpreted as a practical post-hoc linear-decorrelation baseline, not as a controlled comparison in which every component except the basis is jointly optimized.

\section{Coding, Correction, and Reconstruction Details}
\label{app:coding}
\subsection{Bitstream contents and decoding order}
A compressed item contains the primary-latent and hyperlatent rANS streams, the trained model parameters charged to that item, normalization metadata, tensor dimensions, the aligned-channel layout, and, for LD variants, the global transform and per-sample latent means. Decoding restores the hyperlatent first, then the primary latent using either the scale-only or autoregressive prior. LD variants apply $\mathbf{W}^{\top}$ and restore the transmitted means before the shared synthesis network is evaluated. The physical fields are reconstructed and reassembled before residual correction is applied.

\subsection{Error-control wrapper}
The learned codec is optimized for rate and reconstruction error, but the final operating point is accepted only after the residual backend satisfies Eq.~\eqref{eq:nrmse}. For each learned reconstruction, GAE and LBRC are evaluated under the same target tolerance, and the lower-rate valid correction stream is selected. If the learned reconstruction already satisfies the target, the correction rate is zero. Thus, the reported total includes both the compact representation of recurring structure and the sample-specific information required to meet the error criterion. The same rule is applied to CAESAR-V, LD, AR, and LDAR, so the comparison does not favor one learned variant through a different correction policy.

\section{JHTDB Channel Construction}
\label{app:jhtdb}
Our JHTDB experiments use two Cartesian velocity components $(u_x,u_y)$ and pressure $p$ on a three-dimensional grid. 
The shared CAESAR-V backbone processes each input block over time and two in-plane spatial dimensions, so we retain 20 aligned planes along the third spatial dimension.
The internal channel index is therefore the Cartesian product of physical variable and retained plane, giving $G=3\times20=60$ variable--plane channels. LD acts across this aligned channel axis, and AR acts within each two-dimensional latent plane. After inverse transformation and synthesis, the 20 planes are reassembled into the original three-dimensional fields, and Eq.~\eqref{eq:nrmse} is evaluated over the three physical variables. This construction preserves the native extra spatial dimension but means that the JHTDB LD result reflects both cross-variable and cross-plane dependence.

\end{document}